\documentclass[10pt,journal,compsoc]{IEEEtran}

\ifCLASSOPTIONcompsoc
  \usepackage[nocompress]{cite}
\else
  \usepackage{cite}
\fi

\usepackage{subfigure}
\usepackage{algorithm}
\usepackage{algpseudocode}
\usepackage{amsmath}
\usepackage{tabularx}
\usepackage{threeparttable}
\usepackage{makecell}
\usepackage{bbm}
\usepackage{tikz}
\usepackage{pgfplots}
\usepackage{hyperref}
\usepackage[switch]{lineno}
\usepackage{color}
\usepackage{ragged2e}

\begin{document}

\title{Rethinking Natural Adversarial Examples for Classification Models}
\author {
        Xiao Li,
        Jianmin Li,
        Ting Dai,
       	Jie Shi,
       	Jun Zhu,
        Xiaolin Hu \\
\IEEEcompsocitemizethanks{
\IEEEcompsocthanksitem{Xiao Li, Jianmin Li, Jun Zhu and Xiaolin Hu are with the Beijing National Research Center for Information Science and Technology, State Key Laboratory of Intelligent Technology and Systems, Tsinghua University, Beijing 100084, China, and also with the Department of Computer Science and Technology, Tsinghua University, Beijing 100084, China (Corresponding author: Xiaolin Hu.)}
\IEEEcompsocthanksitem{Ting Dai and Jie Shi are with the Trustworthy AI Lab of Shield Lab at Huawei Singapore Research Center.
}
}
}

\newcommand{\tabincell}[2]{\begin{tabular}{@{}#1@{}}#2\end{tabular}}

\IEEEtitleabstractindextext{%
\begin{abstract}
\justifying
Recently, it was found that many real-world examples without intentional modifications can fool machine learning models, and such examples are called ``natural adversarial examples''. ImageNet-A is a famous dataset of natural adversarial examples. By analyzing this dataset, we hypothesized that large, cluttered and/or unusual background is an important reason why the images in this dataset are difficult to be classified. We validated the hypothesis by reducing the background influence in ImageNet-A examples with object detection techniques. Experiments showed that the object detection models with various classification models as backbones obtained much higher accuracy than their corresponding classification models. A detection model based on the classification model EfficientNet-B7 achieved a top-1 accuracy of 53.95\%, surpassing previous state-of-the-art classification models trained on ImageNet, suggesting that accurate localization information can significantly boost the performance of classification models on ImageNet-A. We then manually cropped the objects in images from ImageNet-A and created a new dataset, named ImageNet-A-Plus. A human test on the new dataset showed that the deep learning-based classifiers still performed quite poorly compared with humans. Therefore, the new dataset can be used to study  the robustness of classification models to the internal variance of objects without considering the background disturbance.
\end{abstract}

%\begin{IEEEkeywords}
%Computer Society, IEEE, IEEEtran, journal, \LaTeX, paper, template.
%\end{IEEEkeywords}

}

\maketitle

\IEEEdisplaynontitleabstractindextext

% \fi
\IEEEpeerreviewmaketitle

\IEEEraisesectionheading{\section{Introduction}\label{sec:introduction}}

\IEEEPARstart{T}{he} human visual system is robust not only to subtle changes of an image at the pixel level but also to common image corruptions, such as occlusion, snowflake, and fog scene \cite{geirhos2018generalisation, HendrycksD19}. However, popular deep learning models lack such robustness compared with humans \cite{dodge2017study}. Thus, there is an urgent need to improve the robustness of deep learning models for their wider applications. Adversarial robustness is one of the most popular research topics on the robustness of deep learning models, evaluating the robustness of models facing adversarial examples created by only making some small disturbance (even imperceptible in certain cases) to the pixels of images. Previous work showed that the adversarial examples could make the performance of deep neural network (DNN) classifiers drops dramatically \cite{szegedy2013intriguing}.

 From a broader view, adversarial examples can be ``inputs to machine learning models that an attacker has intentionally designed to cause the model to make a mistake'' \cite{gilmer2018motivating}. In this definition, these adversarial examples include the aforementioned synthetic examples by artificially adding tiny disturbance to the original examples and the real-world, unmodified, and naturally occurring examples \cite{hendrycks2019natural}. The latter adversarial examples are called ``natural adversarial examples'' (NAEs). Such examples are intentionally collected to confuse deep learning models. The ImageNet-A dataset \cite{hendrycks2019natural} is a popular NAEs dataset. On ImageNet-A, the accuracies of ResNets series classifiers trained on the original ImageNet are approximately 5\% \cite{hendrycks2019natural, he2016deep}.

 %Natural adversarial examples also have good black-box transferability, which means various models perform poorly on them.

Adversarial training and data augmentation techniques can significantly improve the performance of classification models on synthetic adversarial examples, however, they have limitations on ImageNet-A \cite{hendrycks2019natural, li2020feature, xie2020adversarial}. The most promising method \cite{xie2020self, mahajan2018exploring} to improve the ImageNet-A accuracy seems to use a large amount of extra data beyond the ImageNet dataset (e.g., 3.5 billion weakly labeled Instagram images) while increase the capacity of deep neural network models (DNNs) (e.g., the large scale version of EfficientNet \cite{tan2019efficientnet}) simultaneously. However, this method is quite time consuming and computationally heavy. Besides, extra data collected from the Internet has the risk of data leakage because they may have a similar origin with the ImageNet-A images.

Although much effort has been put into improving the accuracy of the models on ImageNet-A, most studies did not systematically investigate the reasons for the failure of conventional DNNs on this dataset. Hendrycks et al. \cite{hendrycks2019natural} suggested that the poor performance of the models is mainly due to the occlusion, corruption or unusual angle in the images. By studying this dataset, we hypothesized that large, cluttered and/or unusual background might be another important factor underlying the poor performance. We validated this hypothesis by applying object detection techniques to reduce the influence of background in ImageNet-A examples, and we improved the classification accuracy significantly. Our proposed method achieved the state-of-the-art accuracy among models trained on ImageNet\footnote{Throughout the paper, the ImageNet dataset refers to the dataset used in ILSVRC2012 object recognition task.} \cite{russakovsky2015imagenet}.

\begin{figure*}[!t]
	\centering
 	\includegraphics[width=0.95\linewidth]{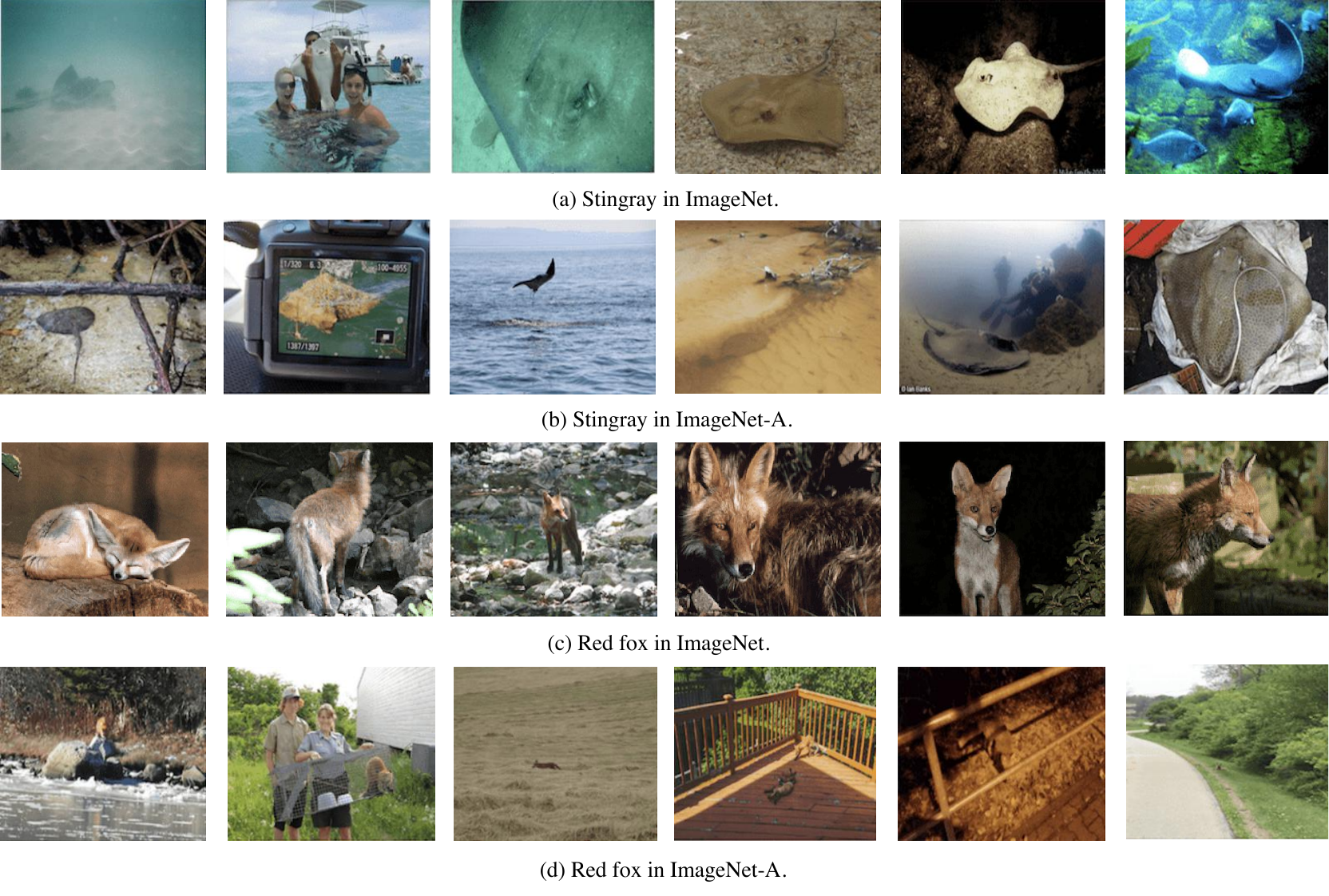}

	\caption{Example images in the \emph{stingray} and \emph{red fox} categories from ImageNet and ImageNet-A.}
	\label{imagenetcom}
\end{figure*}

Object detection results suggested that accurate localization information can significantly boost  the performance of classification models on ImageNet-A. A manually cropping experiment led to the same conclusion. These results suggest that lack of localization mechanism is one of the key reasons for the poor performance of the classification models.

In many scenarios, we want an object classification model which is robust to the internal variation of the objects only. This is the case for object detection as object detection models \cite{ren2015faster, redmon2018yolov3} have other localization mechanisms. The classifier is only used as a backbone to recognize the localized and resized objects. Besides, to study the robustness of classification models, it would be helpful to decouple the influences of the internal variance of the objects (e.g., pose, occlusion and lighting condition) and background disturbance. Thus, by manually removing the background of the target objects in ImageNet-A images, we created a new dataset called ImageNet-A-Plus. Compared with skilled human annotators, existing classification models still performed quite poorly on ImageNet-A-Plus. Nonetheless, the new dataset can be used to study the adversarial robustness of classification models by excluding context interference.

The contributions of this study can be summarized as follows:
\begin{itemize}

    \item We proposed an object detection-based approach to recognize ImageNet-A adversarial examples and achieved significant improvement compared with the classification models, implying that large, cluttered and/or unusual background is one of the important reasons why NAEs in ImageNet-A are difficult to be classified.

    \item We proposed a new dataset called ImageNet-A-Plus, which is publicly available at \url{https://github.com/thu-ml/imagenet-a-plus} to study the adversarial robustness of classification models without considering the background disturbance.

    \item A human experiment showed a large performance gap on robustness between the classification models and the human visual system on ImageNet-A-Plus.
\end{itemize}

\section{Related Work}
\label{sec:relatedwork}

\subsection{Adversarial Attack and Defense}
Since the introduction of adversarial examples, various adversarial attack methods have been developed, e.g., FGSM \cite{goodfellow2014explaining}, PGD \cite{madry2017towards}, Deepfool \cite{moosavi2016deepfool}, BIM \cite{dong2018boosting} and C\&W \cite{carlini2017towards}, and many adversarial defense methods \cite{raghunathan2018certified, DBLP:conf/iclr/TramerKPGBM18} have been designed to tackle these adversarial examples. However, Raghunathan et al. \cite{athalye2018obfuscated} found that most adversarial defense methods relying on obfuscated gradients give a false sense of security, as attackers can evade them through various techniques.

Adversarial training methods \cite{goodfellow2014explaining, deng2020adversarial} are effective against most of the adversarial attacks. A key step in adversarial training is to generate synthetic adversarial examples and use them for training. However, adversarial training has some limitations, such as consuming massive computational resources, taking the risk of decreasing prediction accuracy on normal examples \cite{tsipras2018there} and label leakage \cite{kurakin2016adversarial}. Adversarial training has only slight improvements on NAEs \cite{hendrycks2019natural} because the NAEs are difficult to synthesize.

\subsection{Datasets for Studying Robustness of DNNs}
 Many datasets evaluating the robustness of the ImageNet pretrained models have been proposed. ImageNet-C and ImageNet-P \cite{HendrycksD19} define 15 types of noise, including blur, weather, and different digital corruption types. They are created by modifying the ImageNet validation set with these noises. A brand new ImageNet test set is the ObjectNet \cite{DBLP:conf/nips/BarbuMALWGTK19}, in which the images are obtained by crowd-sourcing with controls over object backgrounds, rotations, and imaging viewpoints.

Unlike the aforementioned datasets, the ImageNet-A \cite{hendrycks2019natural} contains 7500 real-world and unmodified images, which are called ``natural adversarial examples''. The images belong to 200 categories selected from 1000 categories in ImageNet. The dataset was generated by the following steps. First, the images were collected from unmodified original images uploaded by users. Then, a ResNet-50 model pre-trained on ImageNet was used for prediction. If the top-1 prediction of an image was consistent with its user's tag or misclassified with low confidence, it was deleted. Finally, multiple-object images in the selected 200 categories were filtered out manually. The rest of the images are aggregated to be ImageNet-A \cite{hendrycks2019natural}. Compared with other datasets for studying robustness, ImageNet-A is only used for testing, not for training. Mainstream DNN models perform even worse on ImageNet-A \cite{djolonga2020robustness}.

% Several other works \cite{bras2020adversarial, zoran2020towards} reported their ImageNet-A accuracy, but the improvements were still not obvious.

\subsection{General Object Detection}
The object detection task requires a model to output both the locations and the category labels of all target objects in an image. The object detection models can be roughly divided into single-stage methods, e.g., YOLOv3 \cite{redmon2018yolov3} and RetinaNet \cite{lin2017focal} and two-stage methods, e.g., Faster R-CNN \cite{ren2015faster}) and Cascade R-CNN \cite{cai2018cascade}. Techniques such as Region Proposal Network (RPN) \cite{ren2015faster} to pre-extract the objects are often used in two-stage methods but not in single-stage methods. Generally, single-stage methods are faster than two-stage methods, but they suffer from class imbalance and are often inferior to two-stage methods in terms of accuracy. To solve the scale inconsistency problem, Feature Pyramid Network (FPN) \cite{lin2017feature} is used in modern object detection architectures. The FPN leverages different semantic features of DNNs and can be used for detecting objects at different scales.

\section{ImageNet-A Analysis}
\label{sec:analysis}
\subsection{Empirical Observations}
\label{sec:ob}

% We observe obvious data distribution shift between ImageNet and ImageNet-A.
To understand the characteristics of NAEs in ImageNet-A, we studied two categories \emph{stingray} and \emph{red fox} shared in ImageNet and ImageNet-A. Figure \ref{imagenetcom} shows several randomly selected images from these two categories, in which we observed the following.  First, the background of the ImageNet-A images is usually more cluttered, with objects such as digital clock and packet. Second, many target objects occupy a much smaller proportion of images in ImageNet-A compared with those in ImageNet. This is particularly obvious in the red fox category. Third, stingrays in ImageNet are mostly in water, which is their natural living environment. However, some stingrays in ImageNet-A lie in unusual environments such as shore, air and sediment.

By comparing a large amount of examples in the two datasets, we concluded that the aforementioned three phenomena are ubiquitous. In other words, the background of NAEs in ImageNet-A is larger, more cluttered and more unusual than that in ImageNet images. We suspected that this is an important reason why the classification models trained on ImageNet did not perform well on ImageNet-A.

\begin{figure}[!t]
	\centering
	\includegraphics[width=0.95\columnwidth]{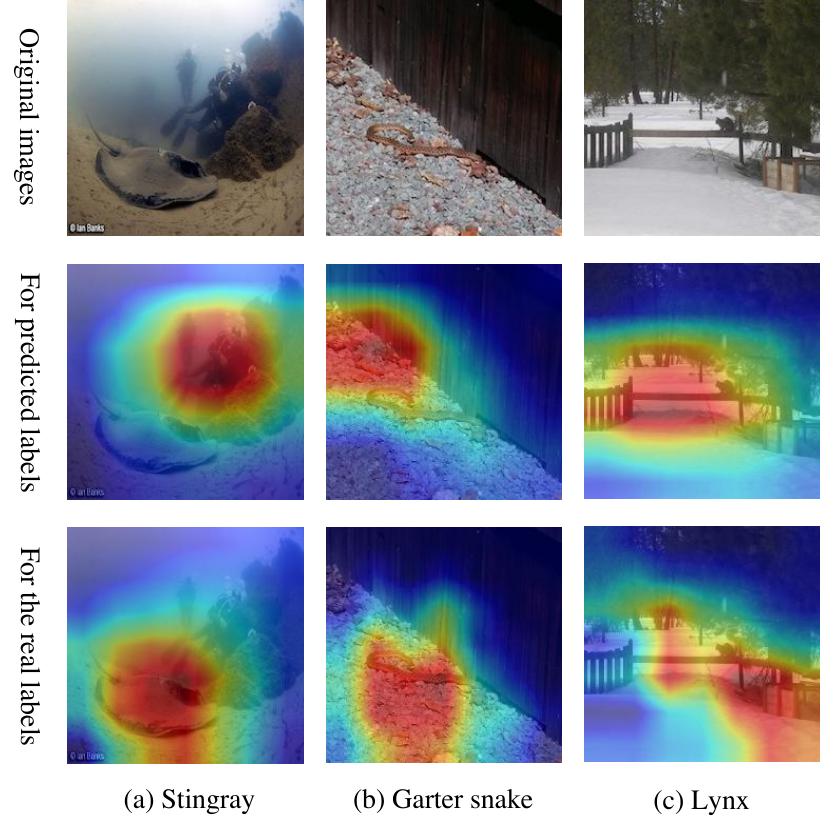}
	\caption{Three example images and their saliency maps generated by Grad-CAM. The model's predictions are (a) \emph{volcano/stingray} with the confidence of 0.878/0.03, (b) \emph{stone wall/garter snake} with the confidence of 0.322/0.0005 and (c) \emph{park bench/lynx} with the confidence of 0.728/0.0007.}
    \label{show}
\end{figure}

\subsection{Explanatory Saliency Map}

Machine learning interpretation tools, such as the explanatory saliency map \cite{sundararajan2017axiomatic, springenberg2014striving}, can be used to analyze the data attribution of ImageNet-A, explaining why the models fail to classify these NAEs. We randomly selected 100 out of the 7500 NAEs from ImageNet-A, and used the Grad-CAM method \cite{selvaraju2017grad} on the ResNeXt50-32x4d model \cite{xie2017aggregated} pretrained on ImageNet to obtain the saliency maps on the images. The saliency map can highlight the important regions for the classification result in an image \cite{selvaraju2017grad}. Figure \ref{show} shows the results of three images. From the saliency maps for predicted labels, the stingray's surrounding background played an important role to misclassify the image as a \emph{volcano} in the second row in Figure \ref{show}(a). Similarly, in the background, the stone ground beside the snake and the wooden bridge under the lynx played an important role in Figure \ref{show}(b) and (c), respectively. If real labels were given, the Grad-CAM method had a big chance to find the correct regions that the model should focus on [third row in Figure \ref{show}(a) and (b)]. Unfortunately, the model's top-1 accuracy was only 6\% on the 100 selected images. Among the misclassified images, approximately 70\% of images had saliency maps for predicted labels whose activation areas were non-overlapped with the positions of the target objects by visual inspection.

From the empirical observation and the saliency map analysis, we hypothesized that the high misclassification rates on ImageNet-A were partly due to the strong background distraction in images. These results suggested that the classification model ResNeXt50-32x4d did not focus on correct regions in the NAEs, which was misled by the background of the objects.

\section{Classification Using Detection Models}
\label{sec:method}

To verify the hypothesis stated in Section \ref{sec:analysis}, a straightforward idea is to crop the object and send it to a classifier. However, this method requires the position of the objects, which is unavailable \emph{ a priori}. Predicting the position of the objects in an image is a subtask of object detection. Therefore, we utilized the localization capability of object detection methods to classify ImageNet-A images.

The challenge of directly applying object detection models on a classification task is that multiple objects belonging to different categories can be detected during the single-image process. Thus, we should consider which categories are the output of the classification task. To overcome the challenge, we converted an object detection model to a classification model by modifying its output in the following ways. First, we kept one object with the highest confidence in each category returned by the object detection model. Second, we sorted the categories by the confidence of the objects. Finally, we outputted the first \emph{K} categories in analogy to the top-\emph{k} outputs of the classifiers. Algorithm \ref{alg:detection} summarizes the entire procedure.

Another problem is the inconsistency between the labels of ImageNet-A and the labels of commonly used datasets for object detection such as MS COCO \cite{lin2014microsoft}. Therefore we cannot pretrain the object detectors on other object detection datasets. Fortunately, some images in the ImageNet dataset have bounding boxes, which are used for the localization task \cite{russakovsky2015imagenet}. We used them to train an object detection model for classification task.

For all the experiments, the classification models were pretrained and fine-tuned on ImageNet, whereas the object detection models were trained as described in Sections \ref{sec:datasetting} and \ref{sec:TP}. Following the literature \cite{hendrycks2019natural, xie2020adversarial}, we only reported the top-1 accuracy.

\begin{algorithm}[t]
  \caption{Classification Based on Object Detectors.}
  \label{alg:detection}
  \begin{algorithmic}
    \Require
    Object detection models with the confidence output $D(*)$, the number of examples $n$, input images $x_1,x_2,\dots,x_n$, collection of categories $C$, $k$ for top-$k$ results.
    \Ensure
   Top-$k$ prediction results $y_1, y_2,\dots, y_k$.

	\For{$i = 1,\dots, n$}
		\State 1. $\{(B_1,c_1, p_1),\dots,(B_k, c_k, p_k),\dots\} = D(x_i)$.
				where $B_x$ denotes the bounding box, $c_x$ denotes the category, $p_x$ denotes the confidence and ``$=$'' denotes the output of $D(x_i)$;
		\State 2. Remove items of $\{(B_x, c_x, p_x)\}$ with the same $c_x$. Only one item with the highest $p_x$ can be reserved per $c_x$;
		\State 3. Sort the rest of $\{(B_x, c_x, p_x)\}$ by $p_x$ from 1 to 0;
		\State 4. Obtain the first $c_1,c_2,\dots, c_k$ in sequence. If the number of $c$ is smaller than $k$, randomly select $c_x$ from $C$;
		\State 5. $y_1, y_2,\dots, y_k = c_1, c_2,\dots, c_k$.
	\EndFor

  \end{algorithmic}
\end{algorithm}
% To be comprehensive, we choose the original classification models as backbone networks of the object detection model.

\subsection{Dataset for Training Object Detectors}
\label{sec:datasetting}

%Each of the 200 ImageNet-A categories has xxx to xxx images (Supplementary Table ?) and in total there are xxx images that can be used to train an object detector.   Note that these images are not well labeled as only one object in most images is labeled with a bounding box though there can be multiple objects in a single image. We believe better performance would be achieved with higher quality of object detection datasets.

Many of the 200 categories of ImageNet-A do not exist in commonly used object detection datasets such as Pascal VOC \cite{everingham2010pascal}, COCO \cite{lin2014microsoft} and Open Image \cite{kuznetsova2018open}. Instead, we used ImageNet data for localization task \cite{russakovsky2015imagenet} to build the training set for our experiment as it contains all categories used in ImageNet-A. Each of the 200 ImageNet-A categories has 211-919 images and there are 110,664 images with bounding box annotations in total. We divided the 110,664 images into training and validation sets in the ratio of 4:1. These images are not well labeled because only one object in most images is labeled with a bounding box, although there can be multiple objects in a single image. We believed that better performance would be achieved with a higher quality of object detection dataset.

\subsection{Implementation Details}
\label{sec:TP}

Although the proposed method is not limited to any object detector, we empirically found that the two-stage model Faster R-CNN \cite{ren2015faster} performed better than several one-stage models \cite{redmon2018yolov3, lin2017focal}. Therefore, we adopted the Faster R-CNN in all experiments, but a better detection model for this task may exist. We used MMDetection \cite{chen2019mmdetection}, an open-source object detection toolbox. We selected ResNet50, ResNeXt101-32x4d, and their versions with SE block \cite{hu2018squeeze} as the backbone networks of Faster R-CNNs.

The model used a five-layer FPN structure. The input channels were 2048, 2048, 1024, 512 and 256 from top to bottom and the output channel was 256. The image resolution was adjusted to 800$\times$600. Data augmentation included random flipping only. The standard stochastic gradient descent (SGD) was used with an initial learning rate of 0.02 and a momentum of 0.9. We trained all Faster R-CNNs for 11 epochs on eight GTX 1080ti GPUs and two images per GPU simultaneously. The learning rate was scaled by 0.1 after the 8th and 10th epochs. We followed the default configurations of Faster R-CNNs to facilitate the training procedure and used the ImageNet pretrained parameter to initialize the backbone network. We set the parameters of the first convolution blocks of backbone networks frozen. The parameters of the remaining stages were fine-tuned together with the parameters of the FPN, RPN, and others.

One can also integrate a very powerful classification model EfficientNet \cite{tan2019efficientnet}, into a detection framework, such as the EfficientDet \cite{Mingxing2019EfficientDet}, for classification. Unfortunately, we could not afford the vast computational resource required to train the EfficientDet. Training a Faster R-CNN with the EfficientNet as the backbone network from scratch was also beyond our capability with our current computing facilities. Alternatively, we modified the internal process of the Faster R-CNN method to conduct the following evaluation:
\begin{itemize}
    \item We used ResNeXt101-32x4d as the backbone network for Faster R-CNN to generate object proposals;

    \item We filtered out proposals that were too small (less than 10 pixels) to be classified or too close to the margin of the image;

    \item We sorted the rest of the proposals by the confidence of being an object and selected the top 20 proposals;

    \item According to Algorithm \ref{alg:detection}, we used the EfficientNet to classify these 20 proposals, and obtained the top-\emph{k} predictions according to the highest confidence of the \emph{softmax} output of the classifier.
\end{itemize}

Here we used the largest and open-source version Efficient-B7 \cite{effipy}. It was trained with AutoAugment \cite{CubukZMVL19}, achieving an accuracy of 84.4\% on the ImageNet dataset. For simplicity, in what follows, we call this method as \emph{Faster R-CNN with EfficientNet as the backbone}, although ResNeXt101-32x4d was used for generating proposals.

\begin{figure}[!t]
	\centering
	\includegraphics[width=0.95\linewidth]{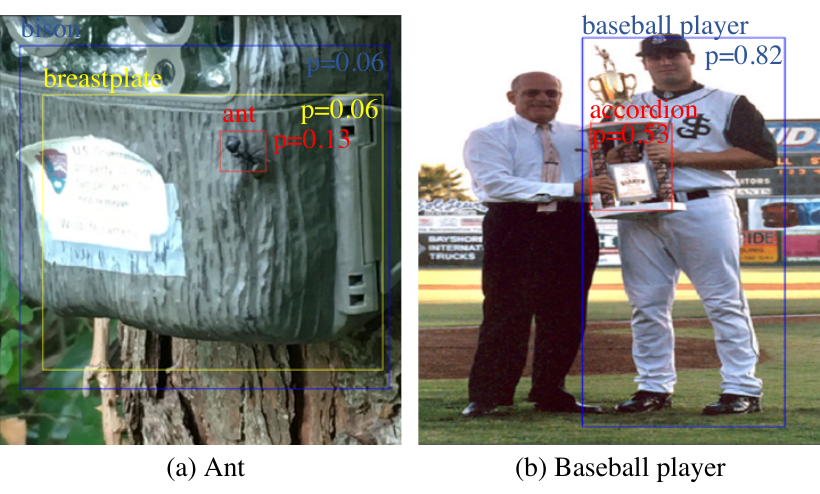}
	\caption{Example results of Faster R-CNN[ResNeXt101-32x4d], showing only the bounding boxes with confidence higher than 0.05.}
    \label{fig:detect}
\end{figure}

\subsection{Detection Models Versus Classification Models}
\label{sec:result}

\begin{table}[!t]
  \centering
  \renewcommand{\arraystretch}{1.3}
  \caption{Comparison between detection models and classification models on ImageNet-A.}
  \begin{minipage}[t]{\linewidth}{\centering}

  \begin{threeparttable}
  	\fontsize{9pt}{\baselineskip}
	\selectfont
    \begin{tabularx}{\linewidth}{cc}
      \Xhline{1.2pt}
      {Method Name} & {Top-1 Acc(\%)}  \\ \Xhline{1pt}

      ResNet50[224$\times$224] & 0.00 \\
      ResNet50[800$\times$600] & 3.91 \\
      Faster R-CNN[ResNet50] & \bfseries 10.23 \mdseries  \\ \hline

      SE-ResNet50[224$\times$224] & 6.17 \\
      SE-ResNet50[800$\times$600] & 7.28\\
      Faster R-CNN[SE-ResNet-50] & \bfseries 10.44 \mdseries \\ \hline

      ResNeXt101-32x4d[224$\times$224] & 5.85 \\
      ResNeXt101-32x4d[800$\times$600] & 7.43 \\
      Faster R-CNN[ResNeXt101-32x4d] & \bfseries 22.31 \mdseries  \\ \hline

      SE-ResNeXt101-32x4d[224$\times$224] & 14.17 \\
      SE-ResNeXt101-32x4d[800$\times$600] & 15.32 \\
      Faster R-CNN[SE-ResNeXt101-32x4d] & \bfseries 22.17 \mdseries \\ \hline

      EfficientNet-B7[600$\times$600] & 38.09 \\
      Faster R-CNN[EfficientNet-B7] &  \bfseries 45.74 \mdseries \\
      \Xhline{1.2pt}
    \end{tabularx}
        \end{threeparttable}
  	\label{tab:result}
   \end{minipage}
\end{table}

Figure \ref{fig:detect} shows some example results of Faster R-CNN[ResNeXt101-32x4d]. Figure \ref{fig:detect}(a) shows that the object detection model detected \emph{ant}, \emph{breastplate}, and \emph{bison} with the confidence of 0.13, 0.06, and 0.06, respectively; thus, the classification result was \emph{ant} by Algorithm \ref{alg:detection}. Similarly, in Figure \ref{fig:detect}(b), the object detection model detected \emph{baseball player} and \emph{accordion} with the confidence of 0.82 and 0.53, respectively; thus, the classification result was a baseball player.

Table \ref{tab:result} shows the performances of several typical classification models and the Faster R-CNNs using the classification models as backbones on ImageNet-A. The brackets following Faster R-CNNs denote the backbone networks. The original classification models except EfficientNet-B7 used an input size of 224$\times$224, whereas Faster R-CNNs used an input size of 800$\times$600. For fair comparison, we fine-tuned all ImageNet pretrained models with 800$\times$600 pixels as input on ImageNet for 10 additional epochs.

As reported in Table \ref{tab:result}, we found the following results. First, the performance of any classification model was boosted by enlarging the input images. Second, sophisticated models, e.g., EfficientNet-B7, tended to perform better than smaller models. Third, the classification models with SE self-attention block \cite{hu2018squeeze} performed better than the original models. Finally, all the object detection models outperformed corresponding classification models.

We also observed that the SE self-attention block improved the classification models but did not improve the object detection models. It might be because both the SE module and object detection techniques can mitigate the impact of background information, and they could not bring new improvement to each other.

\subsection{Comparison with Existing Methods}

\begin{table}[!t]
  \centering
  \renewcommand{\arraystretch}{1.3}
  \caption{Comparison between the proposed detection-based methods on ImageNet-A.}
  \begin{minipage}[t]{\linewidth}{\centering}

  \begin{threeparttable}
  	\fontsize{9pt}{\baselineskip}
	\selectfont
    \begin{tabularx}{\linewidth}{cc}
      \Xhline{1.2pt}
      {Method Name} & {Top-1 Acc(\%)}  \\ \Xhline{1pt}

      ResNet-50 + Cutmix \cite{yun2019cutmix} & 7.30 \\
      ResNet-50 + Cutmix + MoEx \cite{li2020feature} & 8.40 \\
      ResNet-152 + S3TA-16 \cite{zoran2020towards} & 7.83  \\
      Faster R-CNN[ResNet50] & \bfseries 10.23 \mdseries \\ \hline
      EfficientNet-B7 + AdvProp \cite{xie2020adversarial} & 44.70 \\
      Faster R-CNN[EfficientNet-B7 + AdvProp] & \bfseries 53.95 \mdseries \\

      \Xhline{1.2pt}
    \end{tabularx}
%     \begin{tablenotes}
%        \footnotesize
%      \end{tablenotes}
        \end{threeparttable}
  	\label{tab:result2}
   \end{minipage}
\end{table}

We compared our method with several previous methods on ImageNet-A such as \emph{AdvProp} \cite{xie2020adversarial}, \emph{Cutmix} \cite{yun2019cutmix} and \emph{S3TA} \cite{zoran2020towards}. \emph{AdvProp} improves image classification models by using an enhanced adversarial training scheme which treats adversarial examples as additional examples. \emph{Cutmix} performs data augmentation by cutting patches and pasting them on training images, while the ground truth labels are also mixed proportionally to patch area. \emph{S3TA}, standing for a soft, sequential, spatial, top-down attention mechanism, is a regularization strategy to train classifiers with localizable features. It draws inspiration from the primate visual system which is more robust than conventional DNN models.

As shown in Table \ref{tab:result2}, the object detection model with ResNet-50 backbone surpassed different ResNet classification models with various techniques. Note that \emph{AdvProp} alone improved the accuracy of the classification model EfficientNet-B7 by 6.61\%. As an orthogonal approach, we combined our object detection mechanism with \emph{AdvProp} and achieved an even higher accuracy of 53.95\%. This is the best result obtained by models trained on ImageNet to the best of our knowledge\footnote{Note that all models compared in Table \ref{tab:result2} were trained solely on ImageNet. Although \cite{xie2020self} reported the top-1 accuracy of 83.7\% on ImageNet-A, they used 3.5 billion weakly labeled Instagram images as extra data and a larger model EfficientNet-L2.}.

\begin{figure*}[!t]
	\centering
	\includegraphics[width=0.95\linewidth]{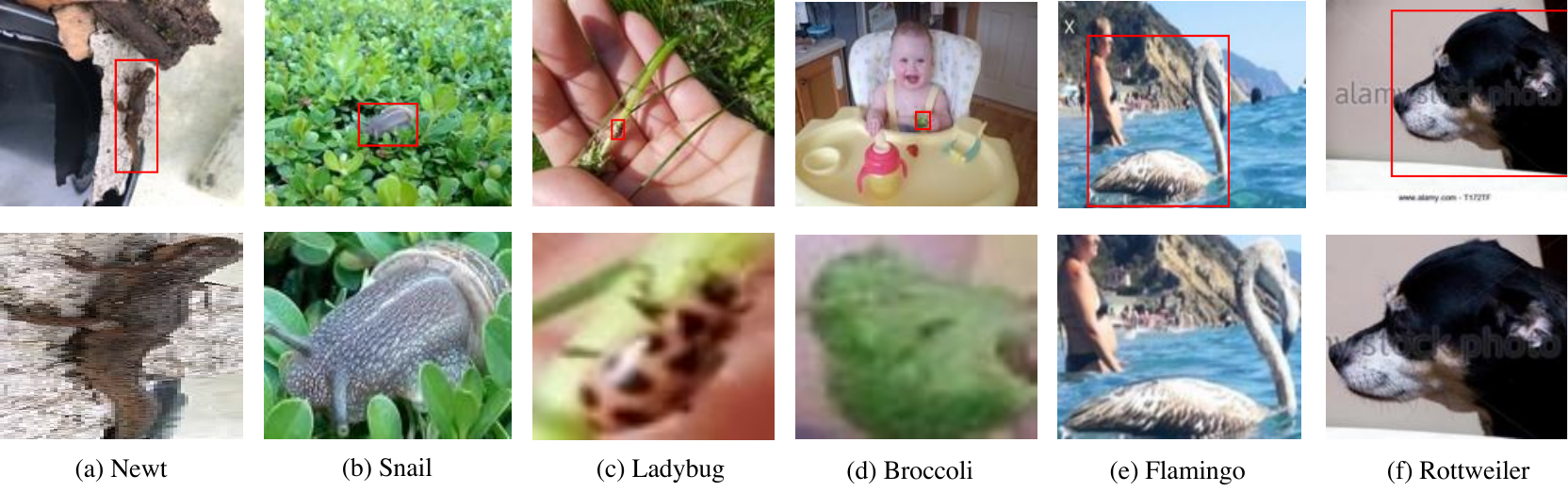}
	\caption{Images before (top row) and after (bottom row) cropping. They are all resized to the same size for display.}
    \label{fig:cropping}
\end{figure*}

\section{Manually Cropped Images}
\label{cropping}

\begin{figure}[!t]
	\centering
	\includegraphics[width=0.95\linewidth]{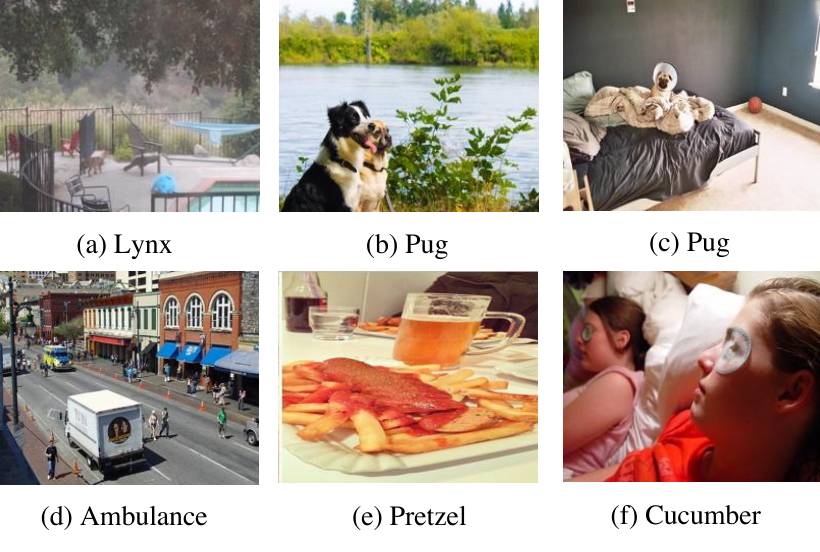}
	\caption{Examples of deleted images from ImageNet-A when constructing the ImageNet-A-Crop dataset.}
    \label{facut}
\end{figure}

\begin{table}[!t]
  \centering
  \renewcommand{\arraystretch}{1.3}
  \caption{The top-1 accuracies before and after cropping with the same input size.}
  \begin{minipage}[t]{\linewidth}{\centering}
  \begin{threeparttable}
  	\fontsize{9pt}{\baselineskip}
	\selectfont
    \begin{tabularx}{\linewidth}{ccc}
      \hline
      {Method Name} & {ImageNet-A} & {ImageNet-A-Crop} \\ \hline
      ResNet50 & 0.00 &  25.17 \\
      ResNeXt50-32x4d & 4.81 &   29.42 \\
      ResNeXt101-32x4d & 5.85 & 34.80 \\
      SE-ResNeXt101-32x4d & 14.17 &  38.03 \\
      EfficientNet-B7 &  38.09 & 49.29 \\
      EfficientNet-B7 + AdvProp & \bfseries 44.73  \mdseries &  \bfseries 59.63 \mdseries \\
      \hline
    \end{tabularx}
        \end{threeparttable}
  \label{tab:cut}
   \end{minipage}
\end{table}

Besides applying object detection models, we also cut out the background manually and performed classification to investigate the influence of the background around the target objects.

We recruited 10 volunteers, each dealing with 20 categories of ImageNet-A, to crop the target objects with rectangular bounding boxes. The labeling method was to select the top, bottom, left and right marginal points of an object and calculate the bounding box. If an image has multiple target objects, the bounding box covers all of them.  We removed a small number of images from ImageNet-A, which have multiple objects of different categories or cannot be recognized by the volunteers, even knowing the labels. Figure \ref{facut} shows some examples we deleted when we constructed the ImageNet-A-Crop dataset. For example,  Figure \ref{facut}(a) is labeled as \emph{lynx}, however, there are chairs in it and the \emph{chair} is another category in ImageNet-A. Figure \ref{facut}(b)(c) also have multiple categories of objects: (b) \emph{pug} and \emph{Rottweiler}; (c) \emph{pug} and \emph{basketball}. Figure \ref{facut}(d) is labeled as an \emph{ambulance} but in fact, there is no ambulance in it. Figure \ref{facut}(e)(f) have similar problems: there is no pretzel among the food in Figure \ref{facut}(e), and the eyeshades in Figure \ref{facut}(f) are not cucumber slices. This procedure resulted in 7373 cropped images, forming a new dataset called \emph{ ImageNet-A-Crop}. Figure \ref{fig:cropping} shows some examples from both ImageNet-A and ImageNet-A-Crop.

We tested the performance of DNNs on these cropped images. All models were pretrained on ImageNet without fine-tuning. The input size for these models was 224$\times$224 except for EfficientNet-B7. To account for the scale inconsistency of objects in the images before and after cropping, we set the input size for EfficientNet-B7 to be 600$\times$600 and 400$\times$400 on ImageNet-A and ImageNet-A-Crop, respectively. Table \ref{tab:cut} shows the prediction results on ImageNet-A and ImageNet-A-Crop \footnote{ImageNet-A-Crop has 127 examples fewer than ImageNet-A, which occupy only 1.69\% of ImageNet-A. Therefore, the difference in the number of examples in the two datasets has little influence on our conclusion.}. Even the networks with poor performance on ImageNet-A, can benefit greatly from the cropping operation. Although EfficientNet-B7 with \emph{AdvProp} is one of the best prediction models on ImageNet-A, manually cropping operation improved its accuracy from 44.73\% to 59.63\%.

We kept the same input resolution before and after cropping, resulting in inevitably enlarged objects after cropping. Thus, the improvement on ImageNet-A-Crop may not come from removing the background alone, but from the enlarged size of the images. The same argument is true for the improvement gained by using the Faster R-CNN technique as the proposals were resized to the same size. We wanted to understand the specific contributions of the two factors.

\begin{figure}[!t]
	\centering
	\includegraphics[width=0.95\linewidth]{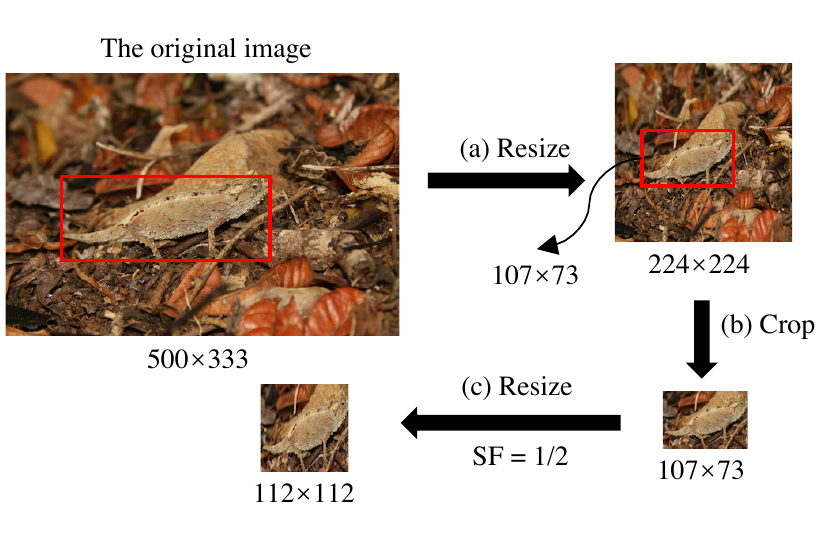}
	\caption{Cropping process for minimizing the influence of the change of image size.}
    \label{demo}
\end{figure}

\begin{table}[!t]
  \centering
  \renewcommand{\arraystretch}{1.3}
  \caption{The top-1 accuracies of the pretrained ResNet-50 model with different scales of cropped images as input.}
  \begin{minipage}[t]{0.8\linewidth}{\centering}
  \begin{threeparttable}
  	\fontsize{9pt}{\baselineskip}
	\selectfont
    \begin{tabularx}{\linewidth}{cccc}
      \hline
      {SF} & {Number} & \tabincell{c}{Accuracy of \\$({\rm SF}\times224)^2$} & \tabincell{c}{Accuracy of \\$({\rm SF}\times448)^2$}  \\ \hline

      1 & 2140 & 11.92 & 13.83 \\
      1/2 & 2213 & 23.77 & 30.95 \\
      1/4 & 2034 & 19.71 & 37.61 \\
      1/8 & 868 & 7.60 & 27.53 \\
      1/16 & 118 & 0.85 & 6.78 \\ \hline
      all & 7373 & 16.94 & 27.03 \\

      \hline
    \end{tabularx}
        \end{threeparttable}
  \label{tab:resolution2}
   \end{minipage}
\end{table}

To do so, ideally, we can crop the target objects from the resized original images and directly use the cropped objects as the input without resizing them again, as the steps (a)(b) shown in Figure \ref{demo}. If a DNN model can obtain higher accuracy on such cropped images (e.g., the image with a size of 107$\times$73) than that on  ImageNet-A with a fixed input size of 224$\times$224, the improvement would mainly come from the operation of background removal, because the target object has the same size in the two cases. But the input size of the ResNet50 model we used cannot be arbitrary. It only accepts input sizes of 224$\times$224, 112$\times$112, 56$\times$56, and so on. Hence, we divided the cropped images into five subsets, and the images were grouped together if they had similar sizes. We used a unified input size for each subset.

Specifically, we used scale factor(SF) to divide the cropped images into different subsets, which can be computed as follows:

$$
{\rm SF} = \sum_{0 \leq k \leq 4}{ 2^{-k} \mathbbm{1}(2^{-k-1} < \max(\frac{w}{224}, \frac{h}{224}) \leq 2^{-k}) }
$$
where $w$ and  $h$ are width and height of the object, respectively, $k$ is an integer and $\mathbbm{1}(*)$ is the indicator function. SF can be $2^{-k}, 0 \leq k \leq 4$.

Figure \ref{demo} shows that the size of the target object is 107$\times$73 when the original image is scaled to 224$\times$224. Thus, the \emph{SF} is computed as $1/2$, and the cropped image is predicted by input size $({\rm SF}\times224)\times({\rm SF}\times224)=112\times112$. This ensures that the step (c) in Figure \ref{demo} has the smallest change to the size of the cropped images.

To predict these cropped images with different input sizes for each subset, we used the pretrained ResNet-50 model without fine-tuning, and the results are shown in Table \ref{tab:resolution2}. For example, 2213 images were predicted with an input size of $112\times112$ and the top-1 accuracy was 11.92\%. Having mitigated the influence of scale, we obtained the top-1 accuracy of 16.94\% among all the 7373 images, which was lower than that tested by a unified input size of 224 × 224 [25.17\% (Table 3)]. Thus, the huge gain obtained by cropping images can come from two aspects. The improvement from 0\% to 16.94\% was mainly because of the reduction of the background interference and the improvement from 16.96\% to 25.17\% was mainly because of the increase of the object scale.

When input size of all images  was enlarged twice, the top-1 accuracy was reached to 27.03\%, implying that further increasing the scale was helpful to recognition. This was especially obvious in recognizing small objects, for example, an absolute improvement of 19.93\% was obtained among the 868 images whose \emph{SF} was $1/8$ when the input size was enlarged twice.

\section{ImageNet-A-Plus}
\label{sec:plus}

\begin{figure*}[!t]
	\centering
	\includegraphics[width=0.95\linewidth]{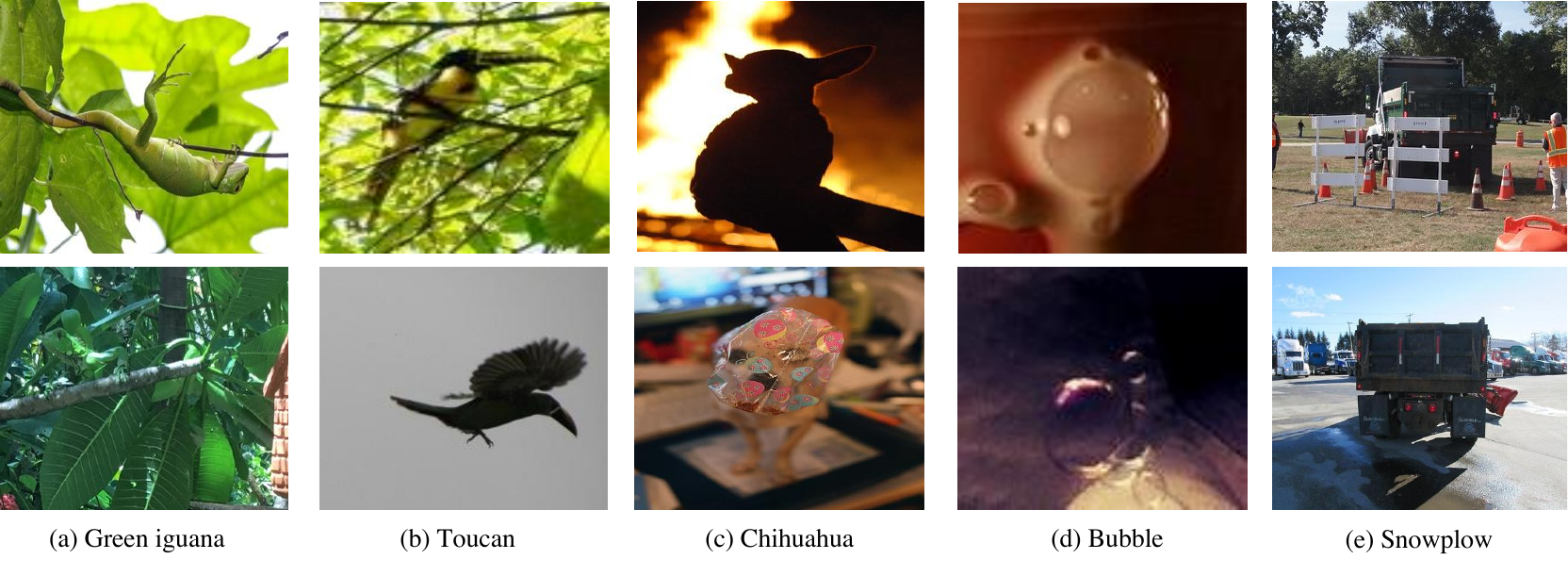}
	\caption{Examples in ImageNet-A-Plus.}
    \label{fig:plus}
\end{figure*}

\subsection{Rethinking NAEs in ImageNet-A}
By either applying object detection techniques or manually removing the background of the main objects in NAEs, we showed that the performances of the classification models on ImageNet-A could be significantly improved. These results highlight the influence of background, which can easily mislead the classifiers. We have empirically observed at least three reasons  (see Section \ref{sec:ob}):

\begin{itemize}
	\item {\it Cluttered Background.} There can be multiple distracting objects in an image. Although the images with multiple objects in the selected 200 categories were filtered out manually during the construction process of ImageNet-A \cite{hendrycks2019natural}, the distracting objects not belonging to the 200 categories, such as Figure \ref{facut}(a)-(c), were kept. The classifiers may predict a category in which the objects look similar to the distracting object.

	\item {\it Unusual Background.} It refers to the background of the target objects in ImageNet-A images which is rarely seen in the ImageNet dataset. Since the classification models are trained on the ImageNet dataset, it is not surprising that they do not perform well on ImageNet-A.

	\item {\it Large Background.} The target object in an image can be quite small compared with the background. In the standard setting of classification, the deep learning models usually take the fixed-size images as input. If an object only takes a tiny part of the image, the resizing operation may make the object too small to be represented in the feature maps in higher layers of a deep learning model. In fact, from our experience, we humans cannot recognize many tiny objects.

\end{itemize}

 Undoubtedly, the images containing multiple objects are inappropriate for probing the robustness of image classification models. An ideal classification model should be able to handle the other two cases, that is, it should be tolerant of the unusual background and size of the objects. However, nowadays, deep learning-based classification models are usually used as basic modules of systems for solving complex computer vision tasks, such as object detection and instance segmentation \cite{he2017mask, SOLO, xie2020polarmask}. In those systems, the classification models are not required to have a high tolerance to the two factors, because the systems have other mechanisms, e.g., proposals of objects in the form of bounding boxes, to deal with these difficulties. This is supported by the higher prediction accuracies of Faster R-CNNs  (Tables \ref{tab:result} and \ref{tab:result2}). Therefore, from a practical point of view, a new dataset for studying the robustness of classifiers to the internal variance of objects other than the influence of context is demanded.

 It would also be easier to identify the cause of classifiers' fragility and then devise more robust models to the internal variance of objects by excluding the influence of the context.

\subsection{Creating ImageNet-A-Plus}

\begin{table}[!t]
  \centering
  \renewcommand{\arraystretch}{1.3}
  \caption{The top-1 accuracies of the 3286 images on ImageNet-A-Plus and ImageNet-A.}
  \begin{minipage}[t]{\linewidth}{\centering}
  \begin{threeparttable}
  	\fontsize{9pt}{\baselineskip}
	\selectfont
    \begin{tabularx}{\linewidth}{ccc}
      \hline
      Method Name & ImageNet-A-Plus & ImageNet-A  \\ \hline

      ResNet50 & 0.00 & 0.00\\
      SE-ResNet50 & 10.47 & 4.90\\
      ResNeXt101-32x4d & 13.39 & 5.23\\
      SE-ResNeXt101-32x4d & 19.72 & 9.59 \\
      EfficinetNet-B7 & 36.33 & 28.61\\
      EfficinetNet-B7+AdvProp & 49.57 & 35.48 \\
      \hline
    \end{tabularx}
        \end{threeparttable}
  \label{tab:plus}
   \end{minipage}
\end{table}

\begin{figure}[!t]
	\centering
	\includegraphics[width=0.9\linewidth]{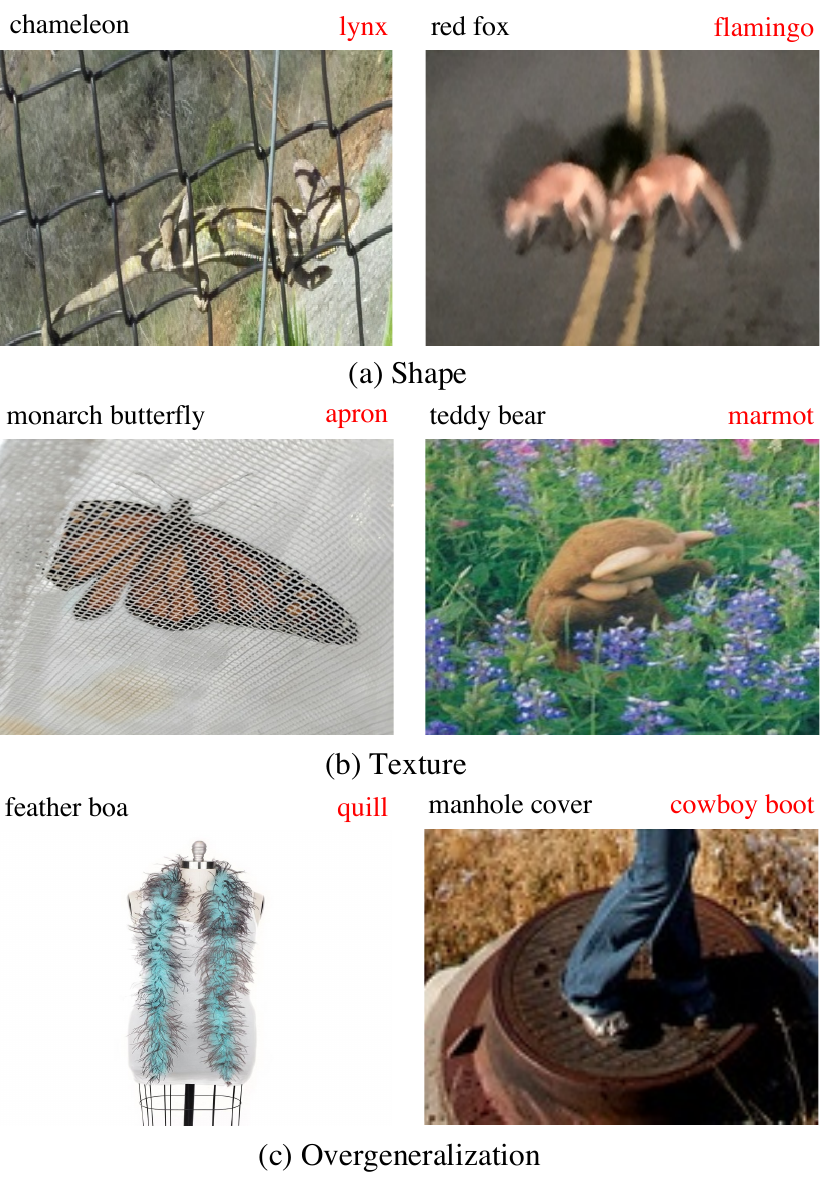}
	\caption{NAEs from ImageNet-A-Plus demonstrating classifier failure modes. Black, the ground truth label. Red, the predicted label.}
    \label{fail}
\end{figure}

From the aforementioned analysis, the images in this new dataset should not contain much background. Instead, they should have only one salient object category, and the proportion of the object should be sufficiently high. We can build this dataset from ImageNet-A.

%**********************

On the basis of the cropping result, for an image in ImageNet-A, we first filtered out it if the proportion of the target object is eight times smaller than the average proportion of the objects in ImageNet per category. Then, we followed Hendrycks et al. \cite{hendrycks2019natural} and filtered out the images if the cropped parts could be recognized by ResNet-50 model. Finally, we clipped part of the background based on the bounding box annotations so that the proportion of the target object in an image was similar to the average proportion in the ImageNet localization dataset per category, e.g., 0.12 and 0.68 for \emph{volleyball} and \emph{red fox}, respectively.  The whole process resulted in 3286 images. We called this dataset as ImageNet-A-Plus, which is publicly available. Some examples in ImageNet-A-Plus are shown in Figure \ref{fig:plus}. The examples in ImageNet-A-Plus are closer to the common inputs for classification.

Table \ref{tab:plus} summarizes the performances of the DNNs on ImageNet-A-Plus. For comparison, we also reported the accuracy on the corresponding 3286 images of ImageNet-A. Most classification models had better performances on ImageNet-A-Plus than those on ImageNet-A. This is expected as the new dataset has reduced the influence of the context. By visual inspection, we also found that the classifiers' failure modes on ImageNet-A-Plus were similar to those on ImageNet-A \cite{hendrycks2019natural}, except for the background cues. Figure \ref{fail} shows some typical failure modes of the ResNet-50 model on ImageNet-A-Plus, which also exist on ImageNet-A. \emph{Chameleon} was predicted as \emph{lynx} and the \emph{red fox} was predicted as \emph{flamingo} , which might be caused by the unusual shapes. The classifiers may also rely too heavily on texture, e.g., misclassifying \emph{monarch butterfly} to be \emph{aporn}  because of the gauze above it. And the classifiers often overgeneralize the visual pattern to incorrect categories, such as associating villus with \emph{quill} and shoes with a \emph{cowboy boot}.

\subsection{Human Test}

\begin{table}[!t]
  \centering
  \renewcommand{\arraystretch}{1.3}
  \caption{Human recognition results of the 600 selected images on ImageNet-A-Plus.}
  \begin{minipage}[t]{0.8\linewidth}{\centering}
  \begin{threeparttable}
  	\fontsize{10pt}{\baselineskip}
	\selectfont
    \begin{tabularx}{\linewidth}{cc}
      \hline
      {Case} & {Number}  \\ \hline

      A1 succeeds, A2 succeeds & 384 \\
      A1 succeeds, A2 fails & 93 \\
      A1 fails, A2 succeeds & 76 \\
      A1 fails, A2 fails & 47 \\

      \hline
    \end{tabularx}
   \end{threeparttable}
  \label{tab:human}
   \end{minipage}
\end{table}

\begin{figure}[!t]
	\centering
	\includegraphics[width=0.9\linewidth]{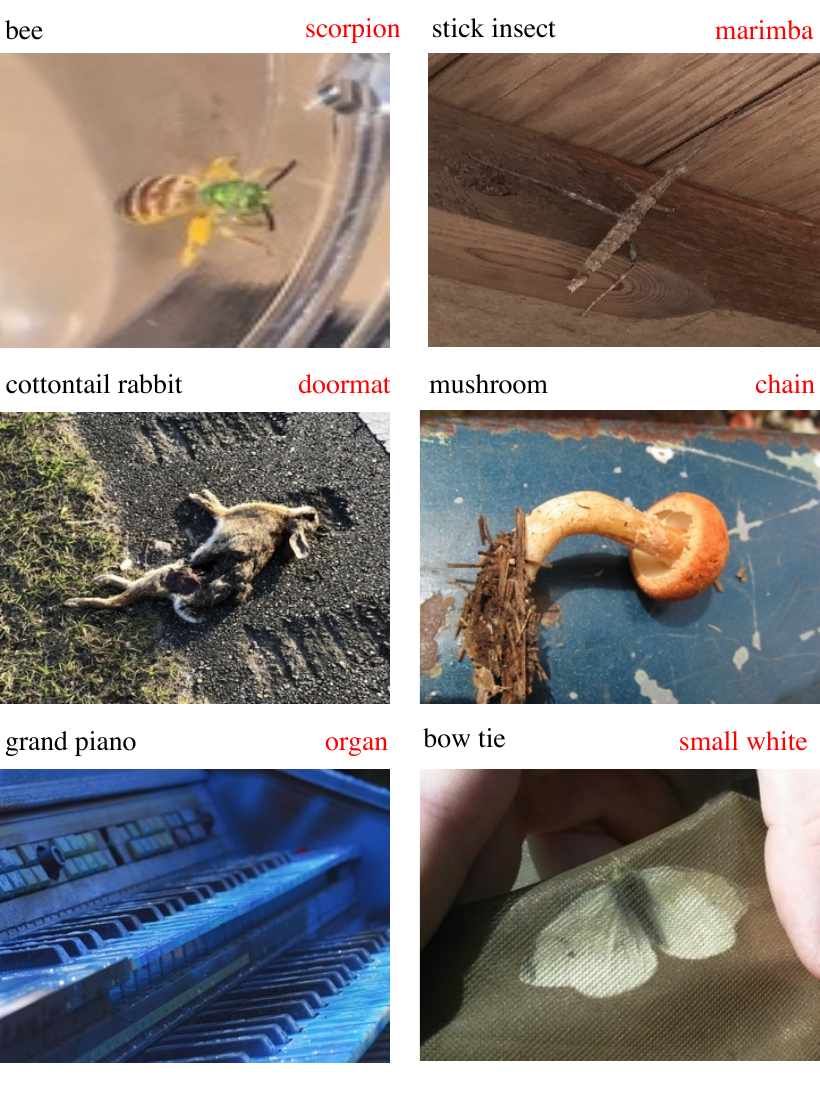}
	\caption{Examples that EfficientNet-B7 with AdvProp failed to recognize. Black, the ground truth label. Red, the predicted label.}
    \label{eff}
\end{figure}

\begin{figure}[!t]
\centering
\fontsize{9pt}{\baselineskip}
\selectfont
\begin{tikzpicture}[scale=0.9]
\begin{axis}[
axis lines=middle,
ymin=5,
%x label style={at={(current axis.right of origin)},anchor=north, below=4mm},
%title={\textit{\textbf{The accuracy of models on different sub-dataset}}},
%xlabel=Subsets,
xticklabel style = {rotate=0,anchor=east, below=1mm},
ylabel=Accuracy(\%),
y label style={at={(current axis.left of origin)},anchor=north, rotate=90, above=8mm, right=20mm},
enlargelimits = false,
xticklabels from table={t.dat}{Subset},xtick=data]
\addplot[orange,mark=square] table [y=SE-ResNet50,x=X]{t.dat};
\addlegendentry{SE-ResNet50}
\addplot[green,mark=star] table [y=ResNeXt101-32x4d,x=X]{t.dat};
\addlegendentry{ResNeXt101-32x4d}]
\addplot[blue,mark=triangle] table [y=SE-ResNeXt101-32x4d,x=X]{t.dat};
\addlegendentry{SE-ResNeXt101-32x4d}]
\addplot[brown,mark=triangle*] table [y=EfficinetNet-B7,x=X]{t.dat};
\addlegendentry{EfficinetNet-B7}]
\addplot[red,mark=square*] table [y=EfficinetNet-B7+AdvProp,x=X]{t.dat};
\addlegendentry{EfficinetNet-B7+AdvProp}]
\end{axis}
\end{tikzpicture}

\caption{The accuracies of models on different subsets of images in ImageNet-A-Plus.}
\label{fig:line}
\end{figure}
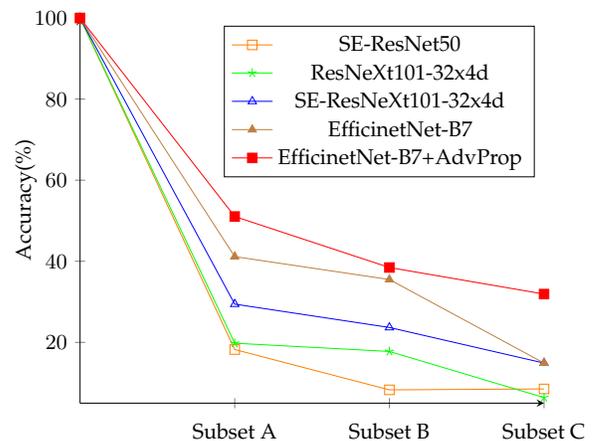

To further study the adversarial robustness on ImageNet-A-Plus, we conducted a human test. We invited four healthy adults (20-24 years of age, two males and two females, with normal or corrected-to-normal vision) to do this experiment. This experiment was approved by the Department of Psychology Ethics Committee, Tsinghua University, Beijing, China.

We split the whole process into training, validation and test steps:

 \begin{itemize}
    \item We randomly selected three images from the ImageNet training set per category in the 200 categories during the training procedure. A total of 600 images were selected. We trained the four annotators on the 600 images.

    \item We validated their performances on the 200 randomly selected images in the ImageNet validation set from the 200 categories and made sure that they could achieve top-1 accuracy of 90\% or higher on the 200 images. Two annotators failed to pass this test, and were excluded from the rest of the experiment.

    \item We tested the performances of the remaining two annotators (A1 and A2) on 600 randomly selected images from ImageNet-A-Plus. The annotators never saw examples in ImageNet-A or ImageNet-A-Plus before. To improve fine-grained recognition accuracy and reduce the influence of human's class unawareness or poor memory, we allowed the annotators to browse the images for training during the test procedure. But in fact, the annotators seldom browsed them (less than 10 times by our monitoring), implying that they were well-trained.
\end{itemize}

Following Russakovsky et al. \cite{russakovsky2015imagenet}, we reported results based on the two experienced annotators, A1 and A2. The results are summarized in Table \ref{tab:human}. A1 and A2 achieved a top-1 accuracy of 79.5\% and 76.7\%, respectively,  which are much higher than those of the pretrained classification models. This gap can be attributed to the internal variance of objects such as the occlusion, corruption or unusual angle in the images, as suggested in the original ImageNet-A study \cite{hendrycks2019natural}.

Figure \ref{eff} shows some examples that EfficientNet-B7 with \emph{AdvProp} \cite{xie2020adversarial}, one of the best DNN models nowadays,  still failed to recognize. These examples are not that hard for humans. The DNNs still need great improvement to achieve human-level robustness on NAEs in ImageNet-A-Plus.

Table \ref{tab:human} suggests that different images posed different levels of difficulty to humans. We were interested in whether these levels aligned with the performances of DNNs. We divided the images into three subsets from easy to hard according to human performance: A, images recognized by both annotators; B, images recognized by one and only one annotator; and C, images not recognized by any annotator. The performances of these DNNs showed a similar trend to humans, as shown in Figure \ref{fig:line} .

\section{Concluding Remarks}
\label{sec:conclusion}
We found that large, cluttered and/or unusual background can have a non-negligible impact on the recognition of NAEs in ImageNet-A. We improved the recognition accuracy by a large margin using object detection models compared with that using the classification models. Considering that most deep learning-based systems for complex computer vision tasks, such as object detection and instance segmentation, have other mechanisms for handling the background problems, we  proposed ImageNet-A-Plus to measure the robustness of classification models to the internal variance of objects without considering the background disturbance. Our human test results on the new dataset showed that the classification models could not compete with humans. Therefore, more efforts are needed to improve the classification models' robustness.

\ifCLASSOPTIONcompsoc
  \section*{Acknowledgments}
\else
  \section*{Acknowledgment}
\fi

This work was supported by the National Key Research and Development Program of China (No. 2017YFA0700904), the National Natural Science Foundation of China (Nos. U19B2034, 62061136001, 61836014 and 61620106010) and a grant from Huawei under contract no YBN2018095341.


\ifCLASSOPTIONcaptionsoff
  \newpage
\fi

\bibliographystyle{IEEEtran}
\bibliography{IEEEabrv,refs}

\begin{IEEEbiography}[{\includegraphics[width=1in,height=1.25in,clip,keepaspectratio]{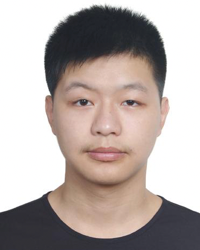}}]{Xiao Li} received the B.E. degree from Tsinghua University, China, in 2020, where he is currently pursuing the Ph.D. degree with the Department of Computer Science and Technology. His research interests are primarily on robustness of machine learning, especially making deep neural networks transparent and robust against adversarial attacks.
\end{IEEEbiography}

\begin{IEEEbiography}[{\includegraphics[width=1in,height=1.25in,clip,keepaspectratio]{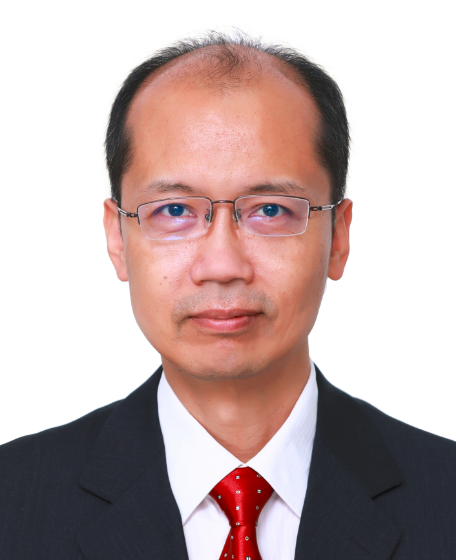}}]{Jianmin Li} received the Ph.D. degree in computer applications from the Department of Computer Science and Technology, Tsinghua University, in 2003. He is currently an Associate Professor with the Department of Computer Science and Technology, Tsinghua University. His main research interests include image and video analysis, image and video retrieval, and machine learning. He has published over 80 journal and conference papers.
\end{IEEEbiography}
\begin{IEEEbiography}[{\includegraphics[width=1in,height=1.25in,clip,keepaspectratio]{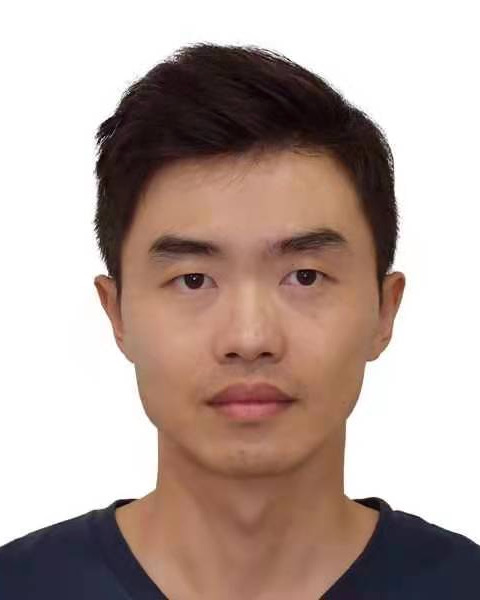}}]{Ting Dai} is a senior researcher in the Trustworthy AI Lab of Shield Lab at Huawei Singapore Research Center. He received his Ph.D. degree in Computer Science from National University of Singapore in 2015, and B.S. degrees in Computer Science from Tsinghua University in 2009. His research area is in system and software security, and security in emerging platforms, such as AI, Web, mobile, and Internet-of-things (IoT). He has been publishing research papers in top security and software engineering conferences and owns several high-value security related patents.
\end{IEEEbiography}

\begin{IEEEbiography}[{\includegraphics[width=1in,height=1.25in,clip]{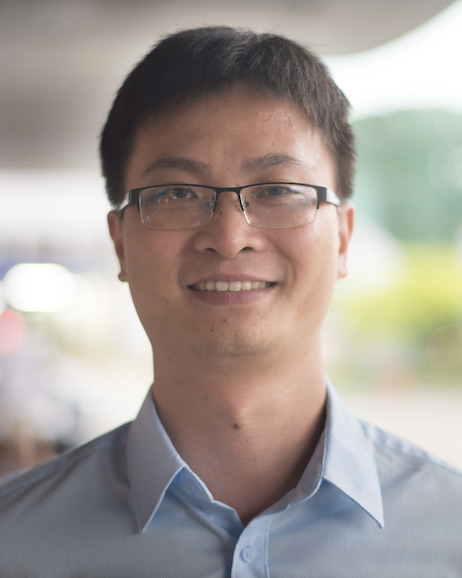}}]{Jie Shi} is a Security Expert in Huawei Singapore Research Center. His research interests include trustworthy AI, machine learning security, data security and privacy, IoT security and applied cryptography. He has over 10 years' research experience, published over 30 research papers in refereed journals and international conferences and obtained over 10 patents. He received his Ph.D degree from Huazhong University of Science and Technology, China.
\end{IEEEbiography}

\begin{IEEEbiography}[{\includegraphics[width=1in,height=1.25in,clip]{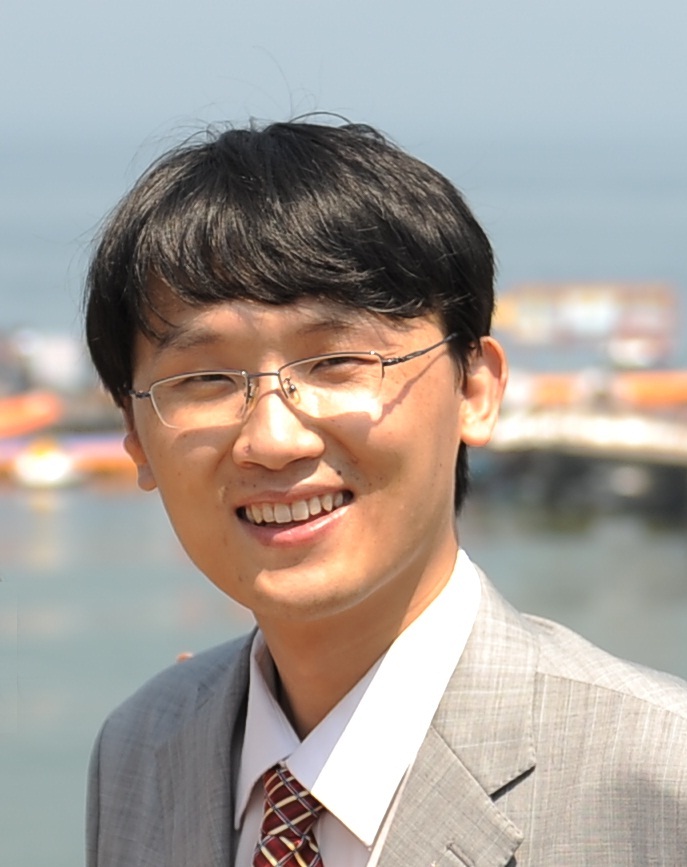}}]{Jun Zhu} received his BS and PhD degrees from the Department of Computer Science and Technology in Tsinghua University, where he is currently a professor. He was an adjunct faculty and postdoctoral fellow in the Machine Learning Department, Carnegie Mellon University. His research interests are primarily on developing statistical machine learning methods to understand scientific and engineering data arising from various fields. He regularly serves as Area Chairs at prestigious conferences, including ICML, NeurIPS, ICLR, IJCAI and AAAI. He is a senior member of the IEEE.
\end{IEEEbiography}
\begin{IEEEbiography}[{\includegraphics[width=1in,height=1.25in,clip,keepaspectratio]{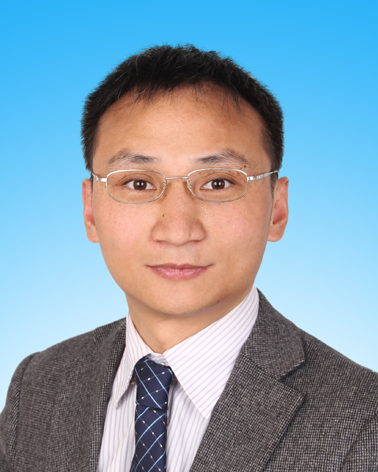}}]{Xiaolin Hu}  (S'01, M'08, SM'13) received the B.E. and M.E. degrees in Automotive Engineering from Wuhan University of Technology, Wuhan, China, and the Ph.D. degree in Automation and Computer-Aided Engineering from The Chinese University of Hong Kong, Hong Kong, China, in 2001, 2004, 2007, respectively. He is now an Associate Professor at the Department of Computer Science and Technology, Tsinghua University, Beijing, China. His current research interests include deep learning and computational neuroscience. He was an Associate Editor of the IEEE Transactions on Neural Networks and Learning Systems. Now he is an Associate Editor of IEEE Transactions on Image Processing and an Associate Editor of Cognitive Neurodynamics.

\end{IEEEbiography}

\end{document}